\documentclass[12pt]{article}
\usepackage{amsmath}
\usepackage{amssymb}
\usepackage{amsthm}
\usepackage[titletoc,toc,title]{appendix}
\usepackage{caption}
\usepackage{caption3}
\usepackage{float}
\usepackage[margin=1in, top = 1in]{geometry}

\usepackage{graphicx}
\usepackage{multirow}
\usepackage{setspace}
\usepackage{subcaption}
\usepackage{threeparttable}
\usepackage[vcentermath]{youngtab}
\numberwithin{equation}{section}

\newtheorem{theorem}{Theorem}

\newenvironment{Proof}[1][Proof]{\noindent\textbf{#1.} }

\begin{document}
\begin{center} \bfseries{Non-negative matrix factorization based on generalized dual divergence} \end{center}
\begin{center}
\bfseries{Karthik Devarajan} \end{center}
\begin{center}
\textit{Department of Biostatistics \& Bioinformatics, Fox Chase Cancer Center, \\ Temple University Health System, Philadelphia, PA} \\
\texttt{karthik.devarajan@fccc.edu}
\end{center}

\begin{center} {\bf Abstract} \end{center}
A theoretical framework for non-negative matrix factorization based on generalized dual Kullback-Leibler divergence, which includes members of the exponential family of models, is proposed.
A family of algorithms is developed using this framework and its convergence proven using the Expectation-Maximization algorithm. The proposed approach generalizes some existing methods for different noise structures
and contrasts with the recently proposed quasi-likelihood approach, thus providing a useful alternative for non-negative matrix factorizations. A measure to evaluate the goodness-of-fit of the resulting factorization is described. This framework can be adapted to include penalty, kernel and discriminant functions as well as tensors.

\section{Kullback-Leibler divergence and its dual}
The {\it Kullback-Leibler} ($KL$) information divergence between two distributions $F$ and $G$ with density (mass)
functions $f$ and $g$ is
\begin{equation}
\label{eq:K}
K (f || g ) \equiv \int \left(\log \frac{f(x)}{g(x)} \right)  d F(x),
\end{equation}
given that $F$ is absolutely continuous with respect to $G$, $F \preceq G$. The discrimination information function in equation (\ref{eq:K}) is a measure commonly used to compare two distributions, and was introduced in Kullback and Leibler (1951). $KL$ information divergence, also referred to as relative entropy or cross-entropy, is the fundamental information measure with many desirable properties for developing probability and statistical methodologies.
Similarly, the measure $K(g||f)$ is known as {\it dual Kullback-Leibler divergence} between $F$ and $G$. In light of the definition above, $K(f||g)$ and $K(g||f)$ are also known as {\it directed divergences}.
These quantities are nonnegative definite and are zero if and only if $f(x) = g(x)$ almost everywhere
(Kullback, 1959; Ebrahimi and Soofi, 2004). One issue pertaining to $K (f || g )$ is that, apart from some exceptional cases such as $F=N(\mu_1, \sigma^2)$ and  $G=N(\mu_2, \sigma^2)$, $K (f || g )$ is not symmetric in $F$ and $G$ where the latter is the reference distribution, i.e., $K (f || g ) \neq K (g || f)$. This lack of symmetry may be of no concern or even desirable in situations where a natural or ideal reference is at hand; e.g., when $G$ is uniform, a natural reference distribution for a problem. However, this is generally not the case for most problems and choice of reference is dependent on the particular application of interest.

Let $\mu_{1}$ and $\mu_{2}$ be the means of random variables corresponding to the probability models $F$ and $G$ with respective densities $f$ and $g$. Then $\beta$-divergence, $D_{\beta}(\mu_{1}||\mu_{2})$, expressed in terms of the means $\mu_{1}$ and $\mu_{2}$ can be written as
\begin{equation}
\label{eq:GenKL}
D_{\beta}(\mu_{1}||\mu_{2}) = \dfrac{1}{\beta(\beta-1)} \left\{{\mu}^{\beta}_{1} - \beta \mu_{1}{\mu}^{\beta-1}_{2} + (\beta - 1)
{\mu}^{\beta}_{2}\right\}, \ \beta \in \Re \backslash \{0,1\}.
\end{equation}
$\beta$-divergence between two densities $f$ and $g$ was introduced by Basu et al. (1998) and Eguchi \& Kano (2001). It has been used by F\'evotte \& Idier (2011) for non-negative matrix factorizations (NMF) where $\beta$-divergence between two objects is considered. In our case, the means $\mu_{1}$ and $\mu_{2}$ represent these objects and we will follow this notation in the remainder of this section. It is well known that $\beta$-divergence in equation (\ref{eq:GenKL}) includes members of the exponential family of models such as the Gaussian $(\beta=2)$, Poisson $(\beta \rightarrow 1)$, gamma $(\beta \rightarrow 0)$ and
inverse Gaussian $(\beta = -1)$ models as special cases. Within this context, $\beta$-divergence can be interpreted as generalized $KL$ divergence indexed by the parameter $\beta$ (Devarajan \& Cheung, 2016). For example, when $\beta=2$ we obtain the Gaussian likelihood $\frac{1}{2}{(\mu_{1}-\mu_{2})}^{2}$, and, in the limit $\beta \rightarrow 0$, we obtain the gamma likelihood $\log \frac{\mu_{1}}{\mu_{2}} - \frac{\mu_{1}}{\mu_{2}} + 1$. In the limit $\beta \rightarrow 1$, we obtain the Poisson likelihood $\mu_{1} \log \frac{\mu_{1}}{\mu_{2}} - \mu_{1} + \mu_{2}$ used in Lee \& Seung (2001). These quantities are commonly referred to as Euclidean distance (ED), Itakuro-Saito (IS) divergence and $KL$ divergence, respectively, in the NMF literature (F\'evotte \& Idier, 2011; Devarajan \& Cheung, 2014; Lee \& Seung, 2001). However, it should be noted that our use of the term $KL$ divergence has a broader connotation similar to that in Devarajan \& Cheung (2014, 2016) and is based on its original definition outlined in Kullback (1951).

We define the generalized dual $KL$ divergence of order $\beta$ by reversing the roles of $\mu_{1}$ and $\mu_{2}$ in equation (\ref{eq:GenKL}). It is given by
\begin{equation}
\label{eq:GendKL}
D^{d}_{\beta}(\mu_{2}||\mu_{1})= \dfrac{1}{\beta(\beta-1)} \left\{{\mu}^{\beta}_{2}-\beta \mu_{2} {\mu}^{\beta-1}_{1} + (\beta-1){\mu}^{\beta}_{1}\right\}, \beta \in \Re \backslash \{0,1\}.
\end{equation}
where the superscript $d$ is used to denote this dual form which also includes, as special cases, members of the exponential family of models as outlined above. When $\beta=2$ we obtain the Gaussian likelihood $\frac{1}{2}{(\mu_{2}-\mu_{1})}^{2}$ which is identical to ED, and, in the limit $\beta \rightarrow 0$, we obtain $-\log \frac{\mu_{2}}{\mu_{1}} + \frac{\mu_{2}}{\mu_{1}} - 1$ which can be viewed as the dual version of IS divergence.
Consider $D^{d}_{\beta}(\mu_{2}||\mu_{1})$ as a function of $\mu_{2}$ with $\mu_{1}$ fixed. Following F\'evotte \& Idier (2011), we find that the first and second derivatives of $D^{d}_{\beta}(\mu_{2}||\mu_{1})$ with respect to $\mu_{2}$ given by
\begin{equation}
\label{eq:D1}
\frac{dD^{d}_{\beta}(\mu_{2}||\mu_{1})}{d\mu_{2}}=\frac{\mu^{\beta-1}_{2}-\mu^{\beta-1}_1}{\beta-1}
\end{equation}
and
\begin{equation}
\label{eq:D2}
\frac{d^{2}D^{d}_{\beta}(\mu_{2}||\mu_{1})}{d\mu^{2}_{2}}=\mu^{\beta-2}_{2},
\end{equation}
respectively, are continuous in $\beta$. It is evident from equations (\ref{eq:D1}) and (\ref{eq:D2}) that $D^{d}_{\beta}(\mu_{2},\mu_{1})$ has a unique minimum at $\mu_{2}=\mu_{1}$ and that it is convex in $\mu_{2}$ for $\beta \in \Re$ (see Figure 1). This contrasts significantly with $\beta$-divergence which is convex in $\mu_{2}$ only for $\beta \in [1,2]$ (F\'evotte \& Idier, 2011).
For a scalar $k > 0$, $D^{d}_{\beta}(\mu_{2}||\mu_{1})$ also satisfies the scale property of $D_{\beta}(\mu_{1}||\mu_{2})$, i.e.,
\begin{equation}
D^{d}_{\beta}(k\mu_{2}||k\mu_{1})={k}^{\beta}D^{d}_{\beta}(\mu_{2}||\mu_{1}).
\end{equation}
Scale invariance is attained for the case $\beta=0$ in equation (\ref{eq:GendKL}) (dual version of IS divergence).

\section{Motivating NMF using generalized dual divergence}
Lee and Seung (1999, 2001) developed NMF algorithms for decomposing a $p \times n$ non-negative matrix $V$ into the product of lower dimensional non-negative matrices $W_{p \times k}$ and $H_{k \times n}$ such that $V \sim WH$, where $k < \frac{np}{n+p}$ is the factorization rank. In order to find an approximation for the input matrix $V$, cost functions that quantify the quality of the approximation need to be constructed using some measure of divergence between $V$ and the reconstructed matrix $WH$. This problem can be formulated in the form of the linear model
\begin{equation}
\label{eq:LM}
V = WH + \epsilon
\end{equation}
where $\epsilon$ is noise. Lee \& Seung's algorithms were based on ED,
\begin{equation}
\label{eq:ED}
L_2(V||WH)=\sum_{ij} {(V_{ij}-(WH)_{ij})}^2,
\end{equation}
and the directed divergence measure,
\begin{equation}
\label{eq:KL}
D(V||WH)=\sum_{ij}\left (V_{ij} \log \frac{V_{ij}}{(WH)_{ij}} - V_{ij} + (WH)_{ij} \right),
\end{equation}
which correspond to the addition of Gaussian and Poisson noise, respectively, in (\ref{eq:LM}).
As noted earlier, the quantity in equation (\ref{eq:ED}) can be derived as $KL$ divergence between two Gaussian random variables with means $\mu_1$ and $\mu_2$ (and equal variance) and the quantity in equation (\ref{eq:KL}) can be derived as $KL$ divergence between two Poisson random variables with means $\mu_1$ and $\mu_2$ (see also Devarajan \& Cheung, 2016).
Unlike $L_2(V||WH)$ which is symmetric, $D(V||WH) \neq D(WH||V)$, so  Lee and Seung (2001)
referred to $D(V||WH)$ as the divergence of $V$ from $WH$. In order to distinguish between the two directed divergences, $D(V||WH)$ and $D(WH||V)$, we use the slight change in notation, $D^{d}(WH||V)$, introduced in equation (\ref{eq:GendKL}). Recently, Devarajan et al. (2015b) derived an algorithm for NMF using the directed divergence $D^{d}(WH||V)$ for the Poisson model given by
\begin{equation}
\label{eq:Pd}
D^{d}(WH||V)=\sum_{ij}\left ((WH)_{ij} \log \frac{(WH)_{ij}}{V_{ij}} - (WH)_{ij} + V_{ij} \right).
\end{equation}
This quantity can be derived as dual $KL$ divergence between two Poisson random variables with means $\mu_1$ and $\mu_2$ as $\beta \rightarrow 1$ in equation (\ref{eq:GendKL}). Similarly, Devarajan \& Cheung (2014) developed NMF algorithms for signal-dependent noise using
\begin{equation}
\label{eq:gammad}
D^{d}(WH||V) = \sum_{i,j} \left\{\log\left(\dfrac{V_{ij}}{{(WH)}_{ij}}\right) + \dfrac{{(WH)}_{ij}}{V_{ij}} - 1\right\}
\end{equation}
for the gamma model and
\begin{equation}
\label{eq:IGd}
D^{d}(WH||V) = \sum_{i,j} \left\{\dfrac{{\left(V_{ij}-{(WH)}_{ij}\right)}^2}{{V_{{ij}}^{2}{(WH)}}_{ij}}\right\}
\end{equation}
for the inverse Gaussian model, quantities that can be derived based on dual $KL$ divergence for the respective models when $\beta \rightarrow 0$ and $\beta=-1$ in equation (\ref{eq:GendKL}).
Furthermore, Dhillon \& Sra (2006) and Cichocki et al. (2009) have proposed NMF algorithms using some special cases of dual divergence.

Since the seminal work of Lee \& Seung (2001), a variety of generalized divergence measures have been utilized for NMF in different applications. Examples include Cheung \& Tresch (2005), Dhillon \& Sra (2006), Kompass (2007), Cichocki et al. (2006, 2008, 2009, 2011), F\'evotte \& Idier (2011) and Devarajan et al. (2015a,b; 2016).
The works of Cheung \& Tresch (2005), Cichocki et al. (2006), F\'evotte \& Idier (2011) and Devarajan \& Cheung (2016) are particularly relevant to the context of this paper.
Cheung \& Tresch (2005) rely directly on the likelihood approach while Cichocki et al. (2006) and F\'evotte \& Idier (2011) utilize $\beta$-divergence in equation (\ref{eq:GenKL}). Recently, Devarajan \& Cheung (2016) proposed a quasi-likelihood approach to NMF based on a unifying theoretical framework using the theory of generalized linear models. It includes all members of the exponential family of models and enables the use of link functions for modeling nonlinear effects. An underlying feature of all these approaches is that they are based on a generalization of $KL$ divergence in some form or another, unified by the approach in Devarajan \& Cheung (2016). Although NMF algorithms for various special cases of generalized dual divergence in (\ref{eq:GendKL}) exist as outlined earlier, a unifying approach that integrates different models and algorithms into a single framework has been lacking.

Within the context of NMF, we can express generalized dual $KL$ divergence of order $\alpha$ between the input matrix $V$ and reconstructed matrix $WH$ as
\begin{equation}
\label{eq:GendKL1}
D^{d}_{\alpha}(WH||V)= \sum_{i=1}^{p} \sum_{j=1}^{n} \frac{\{[(WH)_{ij}]^{2-\alpha}-(2-\alpha)[(WH)_{ij}] V^{1-\alpha}_{ij} + (1-\alpha)V^{2-\alpha}_{ij}\}}{(1-\alpha)(2-\alpha)}, \alpha \in \Re \backslash \{1,2\}.
\end{equation}
using equation (\ref{eq:GendKL}) and the re-parametrization $\beta=2-\alpha$. It is evident from (\ref{eq:GendKL1}) that $D^{d}_{\alpha}(WH||V)$ represents a continuum of divergence measures indexed by the parameter
$\alpha$. More importantly, it embeds the {\it dual KL divergence} of well-known models like the Gaussian ($\alpha=0$), Poisson ($\alpha \rightarrow 1$), gamma ($\alpha \rightarrow 2$) and inverse Gaussian ($\alpha=3$) models. When $1 < \alpha < 2$, it includes the compound Poisson (CP) model which is continuous for $V_{ij} > 0$ but allows exact zeros. By appropriately incorporating a dispersion parameter in (\ref{eq:GendKL1}), $D^{d}_{\alpha}(WH||V)$ includes the quasi-Poisson model which is useful for modeling over- or under-dispersion as $\alpha \rightarrow 1$. Furthermore, it includes the extreme stable ($\alpha \le 0$) and positive stable models ($\alpha > 2$) (Tweedie, 1981; Jorgensen, 1987).

Although $\beta$-divergence includes members of the exponential family of models, it is evident from the work of F\'evotte \& Idier (2011) that a unified NMF algorithm is not feasible due to the non-convexity of the objective function (\ref{eq:GenKL}) for certain ranges of the parameter $\beta$. It turns out that this is not the case with generalized dual $KL$ divergence (\ref{eq:GendKL1}) and that a unified algorithm is indeed possible as shown in the following section. Here, we develop such an algorithm for NMF indexed by the parameter $\alpha$ by minimizing the cost function in equation (\ref{eq:GendKL1}). Such an approach generalizes prior work the work of Devarajan \& Cheung (2014) and Devarajan et al. (2015b) and embeds algorithms for members of the exponential family of models as special cases within a unifying statistical framework.

\section{A unified NMF algorithm based on dual divergence}
We derive a unified NMF algorithm where $\epsilon$ in equation (\ref{eq:LM}) is a member of the class of models included in (\ref{eq:GendKL1}).
One can ignore $\frac{1}{(1-\alpha)(2-\alpha)}$ in (\ref{eq:GendKL1}) and define the function
\begin{equation}
\label{eq:GendKL2}
D^{d}_{\alpha}(WH||V)= \left\{ \begin{array}{l} \displaystyle \sum_{i,j} \left\{[(WH)_{ij}]^{2-\alpha}-(2-\alpha)[(WH)_{ij}] V^{1-\alpha}_{ij} + (1-\alpha)V^{2-\alpha}_{ij}\right\}, \\ \  \  \ \ \ \  \  \  \  \  \  \  \  \  \  \  \  \  \  \  \  \  \  \  \  \  \  \  \  \  \  \  \  \  \  \  \  \  \  \  \  \  \  \  \  \  \  \  \  \  \  \  \  \  \  \  \  \  \  \alpha \in (-\infty,1) \bigcup (2, \infty) \\ \\
\displaystyle \sum_{i,j} \left\{-[(WH)_{ij}]^{2-\alpha}+(2-\alpha)[(WH)_{ij}] V^{1-\alpha}_{ij} - (1-\alpha)V^{2-\alpha}_{ij}\right\}, \\ \  \  \ \ \ \  \  \  \  \  \  \  \  \  \  \  \  \  \  \  \  \  \  \  \  \  \  \  \  \  \  \  \  \  \  \  \  \  \  \  \  \  \  \  \  \  \  \  \  \  \  \  \  \  \  \  \  \  \  \  \  1 < \alpha < 2, \\
\displaystyle \sum_{i,j} \left\{(WH)_{ij} \log\left(\frac{(WH)_{ij}}{V_{ij}}\right) - (WH)_{ij} + V_{ij}\right\}, \alpha=1, \\
\displaystyle \sum_{i,j} \left\{\log\left(\dfrac{V_{ij}}{{(WH)}_{ij}}\right) + \dfrac{{(WH)}_{ij}}{V_{ij}} - 1\right\}, \alpha=2.
\end{array} \right.
\end{equation}
Thus, for any information measure which is proportional to $D^{d}_{\alpha}(WH||V)$ we obtain equation (\ref{eq:GendKL2}). In the case of signal-dependent data such as those observed in various signal processing applications, the divergence in equation (\ref{eq:GendKL2}) offers a flexible choice in decomposing a high-dimensional matrix.

\begin{theorem}
For $\alpha \in \Re \backslash \{1\}$, the measure $D^{d}_{\alpha}(WH||V)$ in equation (\ref{eq:GendKL2}) is non-increasing under the multiplicative update rules for $W$ and $H$ given by
\begin{equation}
\label{eq:H}
H^{t+1}_{aj} = {H}^t_{aj}\left(\dfrac{\sum_i \left({\dfrac{1}{\sum_b W_{ib}H^t_{bj}}}\right)^{\alpha-1} W_{ia}}{\sum_i W_{ia}V^{1-\alpha}_{ij}}\right)^{1/(\alpha-1)}
\end{equation}
and
\begin{equation}
\label{eq:W}
W^{t+1}_{ia} = {W}^t_{ia}\left(\dfrac{\sum_j \left({\dfrac{1}{\sum_b W^t_{ib}H_{bj}}}\right)^{\alpha-1} H_{aj}}{\sum_j H_{aj}V^{1-\alpha}_{ij}}\right)^{1/(\alpha-1)}.
\end{equation}
\noindent This measure is also invariant under these updates if and only if $W$ and $H$ are at a stationary point of the divergence.
\end{theorem}
\begin{Proof}
We provide a more general proof of the monotonicity of updates based on splitting the domain $\Re \backslash \{1\}$ of the parameter $\alpha$ into three disjoint regions and considering them separately.
The update rules for $W$ and $H$ obtained under all cases, however, are the same. A detailed proof of the monotonicity of updates and update rules for the special cases $\alpha=2$ and $\alpha=3$ are provided in Devarajan \& Cheung (2014). In \S3.1, we prove monotonicity of updates and derive update rules for the special case $\alpha=1$.

First, we derive the update for $H$ and prove its monotonicity when $\alpha > 2$ or $\alpha < 1$. Then we show how similar arguments can be used to prove the result for $1 < \alpha < 2$. We will make use of an auxiliary function similar to the one used in the EM algorithm (Dempster et al., 1977; Lee \& Seung, 2001; Devarajan \& Cheung, 2016). Note that for $h$ real, $G(h, h^\prime )$ is an auxiliary function for $F(h)$ if $G(h, h^\prime ) \geq F(h)$ and $G(h, h) = F(h)$ where $G$ and $F$ are scalar valued functions. Also, if $G$ is an auxiliary function, then $F$ is non-increasing under the update $h^{t+1} = \arg \displaystyle\min_h G(h, h^t)$. Using the first equation in (\ref{eq:GendKL2}), we define $$ F(H_{aj}) = (1-\alpha) \sum_i V^{2-\alpha}_{ij} - (2 - \alpha) \sum_{i} \left\{V^{1-\alpha}_{ij} \left(\sum_{a} W_{ia} H_{aj}\right)\right\} + \sum_i \left[ \sum_a W_{ia}H_{aj} \right]^{2 - \alpha},$$ where $H_{aj}$ denotes the ${aj}^{th}$ entry of $H$. Then the auxiliary function for $F(H_{aj})$ is
\begin{eqnarray}
\nonumber
G(H_{aj}, H^t_{aj}) & = & (1-\alpha) \sum_i V^{2-\alpha}_{ij} - (2 - \alpha) \sum_{i} \left\{V^{1-\alpha}_{ij} \left(\sum_{a} W_{ia} H_{aj}\right)\right\} + \\
                    &   &   \sum_{ia} \left\{(W_{ia} H_{aj})^{2-\alpha} \left(\dfrac{W_{ia} H^t_{aj}}{\sum_b W_{ib}H^t_{bj}}\right)^{\alpha -1}\right\}.
\end{eqnarray}
It is straightforward to show that $G(H_{aj},H_{aj}) = F(H_{aj})$. To show that $G(H_{aj},H^t_{aj}) \geq F(H_{aj})$, we use the convexity of $x^{2- \alpha}$ when $\alpha > 2$ or $\alpha < 1$ and the fact that for any convex function $f, f \left( \sum^n_{i=1}r_i x_i \right) \leq \sum^n_{i=1} r_i f(x_i)$ for rational nonnegative numbers $r_1, \cdots , r_n$ such that $\sum^n_{i=1} r_i = 1$. We then obtain
\begin{eqnarray*}
\left( \sum_a W_{ia}H_{aj} \right)^{2 - \alpha} \leq \sum_a \gamma_a \left( \dfrac{W_{ia}H_{aj}}{\gamma_a} \right)^{2 - \alpha} & = &
\sum_a \left\{(W_{ia}H_{aj})^{2 - \alpha} {\left(\dfrac{W_{ia}H^t_{aj}}{\sum_bW_{ib}H^t_{bj}} \right)}^{\alpha-1}\right\},
\end{eqnarray*}
where $\gamma_a = \dfrac{W_{ia} H^t_{aj}}{\sum_b W_{ib}H^t_{bj}}$. From this inequality it follows that $F(H_{aj}) \leq G(H_{aj}, H^t_{aj})$. The minimizer of $F(H_{aj})$ is obtained by solving $$\dfrac{dG(H_{aj},H^t_{aj})}{dH_{aj}} = -(2 - \alpha) \sum_i W_{ia}V^{1-\alpha}_{ij} + (2 - \alpha) \displaystyle \sum_i  \left\{{\left(W_{ia}H_{aj}\right)}^{1-\alpha} W_{ia} \left( \dfrac{W_{ia}H^t_{aj}}{\sum_b W_{ib} H^t_{bj}} \right)^{\alpha -1}\right\} = 0.$$
The update rule for $H$ thus takes the form given in (\ref{eq:H}). For $1 < \alpha < 2$, using the second equation in (\ref{eq:GendKL2}) we define
$$ F(H_{aj}) = -(1-\alpha) \sum_i V^{2-\alpha}_{ij} + (2 - \alpha) \sum_{i} \left\{V^{1-\alpha}_{ij} \sum_{a} W_{ia} H_{aj}\right\} - \sum_i \left[ \sum_a W_{ia}H_{aj} \right]^{2 - \alpha},$$ and the auxiliary function for $F(H_{aj})$ as
\begin{eqnarray}
\nonumber
G(H_{aj}, H^t_{aj}) & = & -(1-\alpha) \sum_i V^{2-\alpha}_{ij} + (2 - \alpha) \sum_{i} \left\{ V^{1-\alpha}_{ij} \sum_{a} W_{ia} H_{aj}\right\} - \\
                    &   & \sum_{ia} \left\{(W_{ia} H_{aj})^{2-\alpha} \left(\dfrac{W_{ia} H^t_{aj}}{\sum_b W_{ib}H^t_{bj}}\right)^{\alpha -1}\right\}.
\end{eqnarray}
It is easy to see that $G(H_{aj},H_{aj}) = F(H_{aj})$. By using the convexity of $-x^{2-\alpha}$ for $1 < \alpha < 2$, we can show that $F(H_{aj}) \leq G(H_{aj}, H^t_{aj})$ and proceed to obtain the update rule for $H$ as described above. The update rule for this case is exactly as that specified for the case $\alpha > 2$ or $\alpha < 1$. By using symmetry of the decomposition $V \sim WH$ and by reversing the arguments on $W$, one can easily obtain the update rule for $W$ given in (\ref{eq:W}) in the same manner as $H$.
\end{Proof}
For a given $\alpha$, we will start with random initial values for $W$ and $H$ and iterate until convergence, i.e, iterate until $|D^{d,(i)}_{\alpha}(WH||V) - D^{d, (i-1)}_{\alpha}(WH||V)| < \delta$ where $\delta$ is a pre-specified threshold between $0$ and $1$ and $i$ denotes iteration number.

\subsection{Special Cases}
As noted before, $D(WH||V)=D(V||WH)=\sum_{ij} {(V_{ij}-(WH)_{ij})}^{2}$ for the Gaussian model corresponding to $\alpha=0$. Hence the NMF algorithm for the Gaussian model based on dual KL divergence is identical o the standard algorithm based on Euclidean distance outlined in Lee \& Seung (2001) (Devarajan \& Cheung, 2014). When $\alpha \rightarrow 2$ and $\alpha=3$ in equation (\ref{eq:GendKL1}), we obtain dual $KL$ divergence for the gamma and inverse Gaussian models in equations (\ref{eq:gammad}) and (\ref{eq:IGd}), respectively. As noted earlier, NMF algorithms for these two models have been described in Devarajan \& Cheung (2014) where monotonicity of updates was proved and update rules were derived for each model. Even though the gamma model is obtained as the limiting case $\alpha \rightarrow 2$ in (\ref{eq:GendKL2}), closed form update rules for $W$ and $H$ can be obtained using $\alpha=2$ in the generalized update rules in equations (\ref{eq:H}) and (\ref{eq:W}). The Poisson special case is discussed below.

\subsubsection{Poisson Model}
When $\alpha \rightarrow 1$ in equation (\ref{eq:GendKL1}), we obtain dual $KL$ divergence for the Poisson model given in equation (\ref{eq:Pd}). Devarajan et al. (2015b) provide an algorithm for this model involving multiplicative updates for $W$ and $H$ but without a formal proof. These update rules are obtained from (\ref{eq:H}) and (\ref{eq:W}) in the limit $\alpha \rightarrow 1$ and are derived in Theorem 2 below.
\begin{theorem}  The measure in equation (\ref{eq:Pd}) is non-increasing under the multiplicative update rules for $W$ and $H$ given by
\begin{equation}
\label{eq:PH}
H^{t+1}_{aj} = {H}^t_{aj} \exp\left(\frac{\sum_{i} W_{ia} \log\left(\frac{V_{ij}}{\sum_b W_{ib}H^t_{bj}}\right)}{\sum_{i}W_{ia}}\right)
\end{equation}
and
\begin{equation}
\label{eq:PW}
W^{t+1}_{ia} = {W}^t_{ia} \exp\left(\frac{\sum_{j} H_{aj} \log\left(\frac{V_{ij}}{\sum_b W^t_{ib}H_{bj}}\right)}{\sum_{j}H_{aj}}\right).
\end{equation}
\noindent This measure is also invariant under these updates if and only if $W$ and $H$ are at a stationary point of the divergence.
\end{theorem}
\begin{Proof}
Using (\ref{eq:H}), the update rule for $H$ for the Poisson model can be written as
\begin{equation}
\label{eq:PH1}
H^{t+1}_{aj} = \lim_{\alpha \rightarrow 1} {H}^t_{aj} \left(\dfrac{\sum_i \left({\dfrac{1}{\sum_b W_{ib}H^t_{bj}}}\right)^{\alpha-1} W_{ia}}{\sum_i W_{ia}V^{1-\alpha}_{ij}}\right)^{1/(\alpha-1)}.
 \end{equation}
The right hand side of (\ref{eq:PH1}) can be re-written as a function of $\alpha$ as
\begin{equation}
\label{eq:PH2}
{H}^t_{aj} \psi(\alpha) = {H}^t_{aj} \left(\dfrac{\sum_i W_{ia}V^{1-\alpha}_{ij}}{\sum_{i} W_{ia}{(\sum_b W_{ib}H^t_{bj})}^{1-\alpha}}\right)^{1/(1-\alpha)}.
\end{equation}
Using (\ref{eq:PH2}) in (\ref{eq:PH1}) and taking logarithm on both sides, we get
$$\log H^{t+1}_{aj} = \log {H}^t_{aj} + \lim_{\alpha \rightarrow 1} \log \psi(\alpha) = \log {H}^t_{aj} +
\lim_{\alpha \rightarrow 1} \frac{1}{1-\alpha} \left\{\log \left(\dfrac{\sum_i W_{ia}V^{1-\alpha}_{ij}}{\sum_{i} W_{ia}{(\sum_b W_{ib}H^t_{bj})}^{1-\alpha}}\right)\right\}.$$ Applying l'Hospital's rule to compute the limit, we obtain
$$\log H^{t+1}_{aj} = \log {H}^t_{aj} + \left(\frac{\sum_{i} W_{ia} \log\left(\frac{V_{ij}}{\sum_b W_{ib}H^t_{bj}}\right)}{\sum_{i}W_{ia}}\right).$$ Hence
\begin{equation}
H^{t+1}_{aj} = {H}^t_{aj} \exp\left(\frac{\sum_{i} W_{ia} \log\left(\frac{V_{ij}}{\sum_b W_{ib}H^t_{bj}}\right)}{\sum_{i}W_{ia}}\right).
\end{equation}
Similarly, the update rule for $W$ can be obtained as specified in (\ref{eq:PW}). Monotonicity of these updates follows directly from the monotonicity of generalized updates in equations (\ref{eq:H}) and (\ref{eq:W}) established in Theorem 1 when $\alpha \rightarrow 1$.
\end{Proof}

\section{Measuring Goodness-of-fit}
The updates derived in equations (\ref{eq:H}), (\ref{eq:W}), (\ref{eq:PH}) and (\ref{eq:PW}) ensure monotonicity of updates for a given run of the NMF algorithm for pre-specified $\alpha$ and rank $r$, based on random initial values for $W$ and $H$. However, NMF algorithms are typically prone to the problem of local minima and, thus, require the algorithm using multiple random restarts. The factorization from the run that produces the best reconstruction, quantified by minimum reconstruction error across multiple runs, can be used for assessing goodness-of-fit. Following Devarajan \& Cheung (2014, 2016), we propose a unified measure for this purpose based on model-specific minimum reconstruction error, $RE$. It quantifies the variation explained by the continuum of statistical models contained in equation (\ref{eq:GendKL2}). For a given rank $r$ the proportion of explained variation, $R^2$, is dependent on the particular model, determined by $\alpha$, used in the factorization and is computed as
\begin{eqnarray} \label{eq:R2}
%\nonumber
R^{2} & = & 1 - \frac{\min D^{d}_{\alpha}(WH||V)}{D^{d}_{\alpha}(\bar{V}||V)}
\end{eqnarray}
where $RE$ is the numerator on the right hand side of equation (\ref{eq:R2}), $D^{d}(WH||V)$ is as specified in equation (\ref{eq:GendKL2}) and $WH$ represents the reconstructed matrix. For rank $r$, the $(i,j)^{th}$ entry of $WH$ is  $(WH)_{ij}=\sum_{a=1}^{r} W_{ia}H_{aj}$; in the denominator, each entry is replaced by the grand mean of all entries of the input matrix $V$, $\bar{V} = \dfrac{1}{np} \left\{\sum_{i=1}^{p} \sum_{j=1}^{n} V_{ij}\right\}$. Note that when $\alpha=0$, these quantities can be interpreted as the residual and total sum of squares, respectively, associated with the Gaussian model. For the nonlinear models indexed by $\alpha$ in equation (\ref{eq:GendKL2}), $R^2$ measures the proportion of empirical uncertainty explained by the inclusion of $W$ and $H$ (Cameron \& Windmeijer, 1997; Devarajan \& Cheung, 2014; 2016).

\section{Applications}
Several special cases of the proposed unifying framework have been utilized for NMFs involving a variety of applications. For instance, Devarajan \& Cheung (2014) derived algorithms based on dual divergence for gamma and inverse Gaussian models - using equations (\ref{eq:gammad}) and (\ref{eq:IGd}), respectively - for handling signal-dependent noise structures and demonstrated their application in electromyography studies for extraction of muscle synergies. These methods explained more variation ($R^2$) in the data at the appropriate number of synergies identified for each data set in a study involving frog motor behaviors under different experimental conditions. Similarly, Devarajan et al. (2015b) proposed an algorithm for the Poisson model based on dual divergence in equation (\ref{eq:Pd}) for unsupervised dimension reduction of discrete multivariate data. Two benchmark data sets - the Reuters news groups data and the Sacchromyces Genome Database (Shahnaz et al., 2006; Chagoyen et al., 2006) - were utilized for this purpose. In both cases, the algorithm based on dual divergence resulted in the best reconstruction compared to other competing methods. The proposed approach consolidates the above methods as well as a spectrum of other methods into a unifying framework and, thus, provides a flexible alternative for exploratory analysis of high dimensional data generated by diverse mechanisms that are exclusive to different applications.

\section{Conclusions} \label{CONLC}
In summary, this paper presented a unified approach to NMF based on generalized dual $KL$ divergence along with a rigorous proof of convergence. The proposed approach is broadly applicable to the exponential family of models and is particularly useful in applications where there is {\it a priori} knowledge or empirical evidence of signal-dependence in noise. Furthermore, it unifies various existing algorithms and contrasts with the recently proposed quasi-likelihood approach, thus providing a complementary view of NMF. The basic principle underlying this framework is broadly extensible to the use of penalty, kernel and discriminant functions and to tensors.

\begin{figure} \vspace*{.15in}
\begin{center}
\vspace*{-.15in}
\includegraphics[width=6.2in,height=5in]{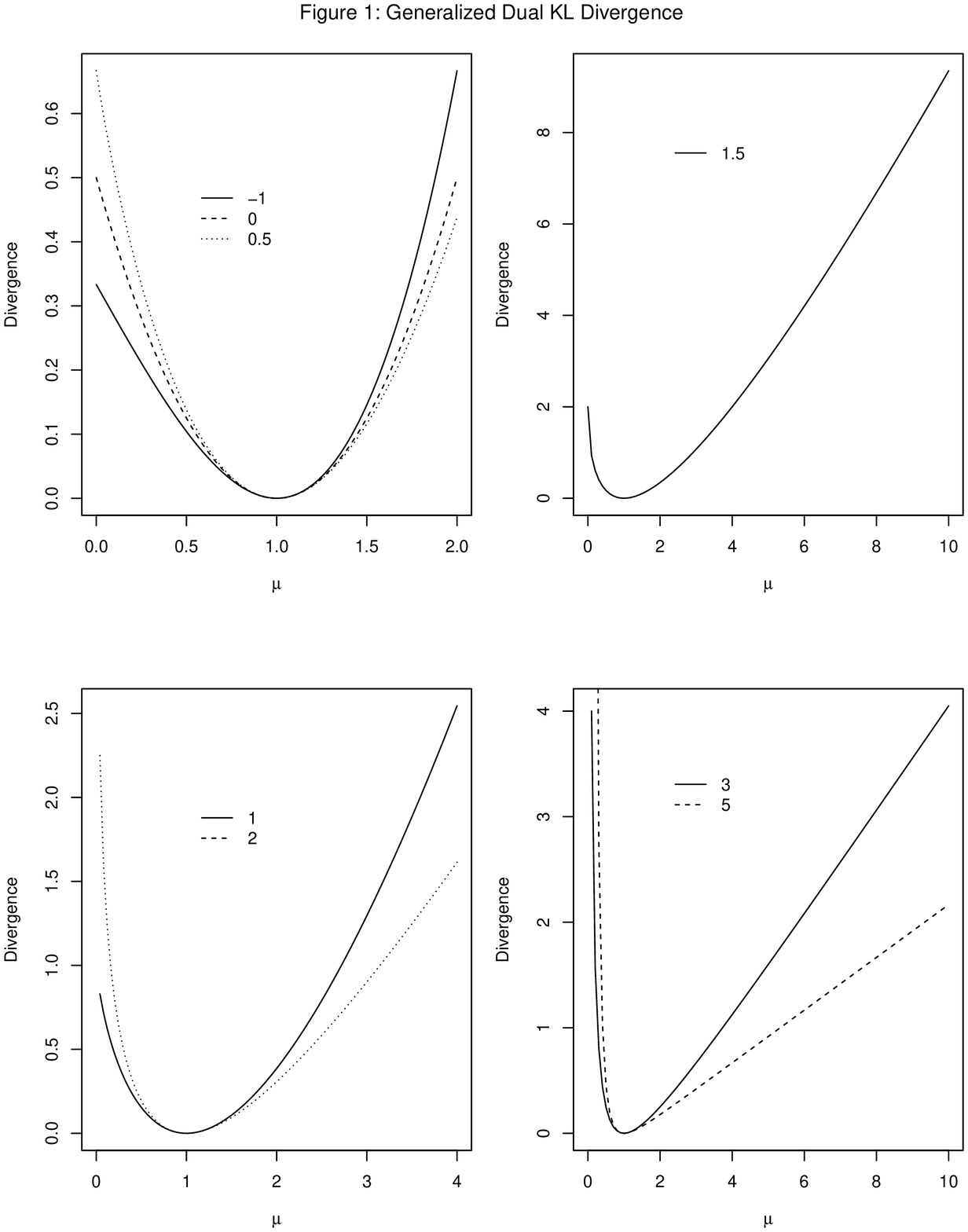}
\end{center}
\end{figure}

\section*{Figure Legend}
Figure 1, panels (a)-(d): Generalized dual $KL$ divergence, equation (\ref{eq:GendKL}), plotted as a function of $\mu_{2}=\mu$ for $\mu_{1}=1$ and various choices of $\alpha$, illustrating its convexity across the entire range of $\alpha$. The values of $\alpha$ are indicated in the legend within each panel.

\section*{Acknowledgements}
Research of the author was supported in part by NIH Grant P30 CA06927.

\singlespacing

\end{document}